\documentclass[final,5p,times,twocolumn]{elsarticle}

\usepackage{graphicx}
\usepackage{verbatim}
\usepackage{MnSymbol}
\usepackage{stmaryrd}
\usepackage{hyperref}
\usepackage{amsmath}
\usepackage{subfig}
\usepackage[ruled,vlined]{algorithm2e}

\newcommand{\bs}[1]{\boldsymbol{#1}}
\newcommand{\mc}[1]{\mathcal{#1}}

\renewcommand{\exp}{\textbf{e}}
\newcommand{\dotprod}[1]{\langle \bs \theta, \  \featvec{#1} \rangle}
\newcommand{\dotprodexp}[1]{\sum\limits_{m \in \mc A_k(y_{#1-1}, y_#1)} \theta_m \cdot
\feat{m}{#1}}

\newcommand{\factor}[1]{\exp^{\, \dotprod{#1}}}
\newcommand{\featvec}[1]{\bs f (y_{#1 - 1}, y_#1, \bs x, #1) }

\newcommand{\feat}[2]{f_#1 (y_{#2 - 1}, y_#2, \bs x, #2) }

\newcommand{\p}{p(\bs y | \bs x; \bs \theta)}
\newcommand{\totsum}{\sum_{\y}}

\newcommand{\y}{\bs y}

\newcommand{\Z}{Z(\bs x ; \bs \theta)}
\newcommand{\partialz}[1]{\nabla_{\theta_#1}\Z}
\newcommand{\gradz}{\nabla_{\theta}\Z}

\newcommand{\psimat}[1]{\psi_#1 (y_{#1}, y_{#1-1})}
\newcommand{\umat}[1]{u_#1 (y_{#1-1}, y_{#1})}
\newcommand{\featurevec}[1]{\bs f (y_{#1-1} , y_{#1}, \bs x, #1)}
\newcommand{\tableskip}{\hskip 0 cm}
\newtheorem{definition}{Definition}
\newtheorem{lemma}{Lemma}

\newcommand{\sR}{\mathbb R}
\newcommand{\osum}{\bigovert}

\newcommand{\logp}{\oplus}
\newcommand{\logt}{\otimes}
\newcommand{\logsum}[2]{\bigoplus_{#1}^{#2}}
\newcommand{\logprod}[2]{\bigotimes_{#1}^{#2}}
\newcommand{\logtotsum}{\logsum{\y}{} \ }

\newcommand{\esrp}{\overt}
\newcommand{\esrt}{\odot}
\newcommand{\esrsum}[2]{\bigovert_{#1}^{#2}}
\newcommand{\esrprod}[2]{\bigodot_{#1}^{#2}}

\renewcommand{\log}{\ln}

\begin{document}

\begin{frontmatter}

\title{Gradient Computation In Linear-Chain Conditional Random Fields \\ Using The Entropy Message Passing Algorithm}



\tnotetext[t1]{Research supported by Ministry  of Science and
Technological Development, Republic of Serbia, Grants No. 174013
and 174026}

\author[misanu,fsmun]{Velimir M. Ili\'c\corref{cor}}
\ead{velimir.ilic@gmail.com}

\author[fsmun]{Dejan I. Man\v cev}
\ead{dejan.mancev@pmf.edu.rs}

\author[fsmun]{Branimir T. Todorovi\'c}
\ead{branimirtodorovic@yahoo.com}

\author[fosun]{Miomir S. Stankovi\'c}
\ead{miromir.stankovic@gmail.com}

\cortext[cor]{Corresponding author. Tel.: +381112630170; fax:
+381112186105.}

\address[misanu]{Mathematical Institute of the Serbian Academy of Sciences and Arts, Kneza Mihaila 36, 11000 Beograd, Serbia}
\address[fsmun]{University of Ni\v s, Faculty of Sciences and Mathematics, Vi\v segradska 33, 18000 Ni\v s, Serbia}
\address[fosun]{University of Ni\v s, Faculty of Occupational Safety, \v Carnojevi\'ca 10a, 18000 Ni\v s, Serbia}

%
\begin{abstract}
The paper proposes a numerically stable recursive algorithm for the exact computation of the linear-chain conditional random field gradient. It operates as a forward algorithm over the log-domain expectation semiring and has the purpose of enhancing memory efficiency when applied to long observation sequences. Unlike the traditional algorithm based on the forward-backward recursions, the memory complexity of our algorithm does not depend on the sequence length. The experiments on real data show that it can be useful for the problems which deal with long sequences.
\end{abstract}



\end{frontmatter}

\section{Introduction}

Conditional random fields (\emph{CRF}s) \cite{Lafferty_et_al_01}
are probabilistic discriminative classifiers which can be applied
for labeling and segmenting sequential data. When compared with
more traditional sequence labeling tools like hidden Markov models
(\emph{HMM}s), the \emph{CRF}s offer the advantage by relaxing the
strong independence assumptions required by \emph{HMM}s.
Additionally, \emph{CRF}s avoid the label bias problem
\cite{Lafferty_et_al_01} exhibited by the maximum entropy Markov models
and other conditional Markov models based on directed graphical
models. However, these improvements are accompanied by a significant
cost in time and space needed for the parameter estimation of
\emph{CRF}, especially for real-time problems like labeling very
long sequences which appear in computer security \cite{Lane_00},
\cite{Warrender_et_al_99}, bioinformatics \cite{Krogh_et_al_94},
\cite{Mayer_Durbin_04} and robot navigation systems
\cite{Koening_Simmons_96}.

The \emph{CRF} parameter estimation is typically performed by some of
the gradient methods, such as iterative scaling, conjugate
gradient, or limited memory quasi-Newton methods \cite{Gupta},
\cite{Lafferty_et_al_01}, \cite{Sha_Pereira_03}, \cite{Sutton_08},
\cite{Vishwanathan_et_al_06}. All these methods require the
computation of the likelihood gradient, which becomes
computationally demanding as the sequence length and the number of
classes increase. The standard method for gradient computation
\cite{Lafferty_et_al_01} is based on the internal computation of
\emph{CRF} marginal probabilities by use of
the forward-backward (\emph{FB}) algorithm. %

The \emph{FB} algorithm first appeared in two independent
publications \cite{Baum_Petrie_66}, \cite{Chang_Hancock_66}, but
it is better known from the subsequent papers
\cite{Bahl_et_al_74}, \cite{Baum_72}. It makes use of dynamic
programming, running with the asymptotical time complexity $\mc O
(N^2 T)$ and with the memory complexity $\mc O (N T)$, where $T$
denotes the sequence length and $N$ denotes the number of states
classified. In spite of the time efficiency, it becomes spatially
demanding when the sequence length is exceptionally large
\cite{Khreich_et_al_10}.
Furthermore, when it is used for the linear-chain gradient
computation as in \cite{Gupta}, \cite{Lafferty_et_al_01},
\cite{Sha_Pereira_03}, \cite{Sutton_08},
\cite{Vishwanathan_et_al_06} it requires the storage of all
\emph{CRF} transition matrices which increase the total memory
complexity for $\mc O(N^2 T)$.

The memory complexity can be reduced with modifications of the
\emph{FB} algorithm such as the checkpointing algorithm
\cite{Grice_et_al_97}, \cite{Tarnas_et_al_98} or with the
re-computation of the transition matrices every time they are used
(see section 3.3.). However, these techniques increase the
computational complexity, while the memory complexity still
depends on the sequence length.
Another possibility is the use of forward-only algorithm 
\cite{Churbanov_Winters-Hilt_08},
\cite{Miklos_Meyer_05}, \cite{Sivaprakasam_Shanmugan_95}, for
which the matrices can be computed in runtime. This algorithm runs
with constant memory complexity but it is computationally
inefficient since it runs with the computational complexity $\mc
O(N^4 T)$.


In this paper we propose an algorithm for the exact computation of the
linear-chain conditional random field gradient. The algorithm is derived as a forward algorithm over the introduced log-domain expectation semiring, which means that its recursive
equations can be obtained if real sums and products
from an ordinary FB are replaced with products and sums from the
log-expectation semiring.
Accordingly, it can be seen as a numerically stable version
of our previously developed Entropy Message Passing algorithm (EMP)
\cite{Ilic_et_al}, and it will also be called the EMP. Unlike the standard
procedure, the EMP does not compute each marginal separately, but
computes the gradient in a single forward pass by use of double
recursion.
Since only the forward pass is needed, the EMP can be implemented with
the memory complexity being independent of the sequence length, having
the advantage over the FB when long sequences are used.

The paper is organized as follows: In section II we explain the
\emph{FB} algorithm which operates over a commutative semiring.
In section III we introduce the problem of efficient computation
of a linear-chain \emph{CRF} gradient and review the standard method
based on the \emph{FB} algorithm. The algorithms based on the
\emph{EMP} are presented in section IV, where the complexity
analysis is given. Finally, the experimental results are presented
in section V where two methods are compared and the advantage of
the EMP is discussed.


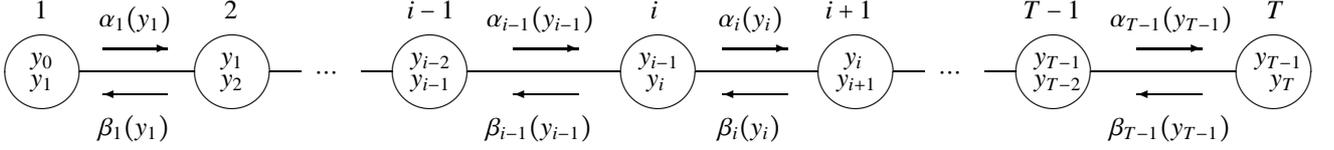
\begin{figure*}[!t]
  \label{JTA_chain} \centering
\setlength{\unitlength}{1mm}
\begin{picture}(170, 15)

  \put(4,16){$1$}

  \put(5, 9){\circle{10}}
  \put(3.5, 10){$y_{0}$}
  \put(3.5, 7){$y_{1}$}


  \put(10,9){\line(1,0){15}}

  \put(13, 12){\vector(1,0){8.5}}
  \put(12.5, 15){$\alpha_{1}(y_1)$}

  \put(21.5, 6){\vector(-1,0){8.5}}
  \put(12.5, 0.5){$\beta_{1}(y_1)$}

  \put(29,16){$2$}

  \put(30, 9){\circle{10}}
  \put(28.5, 10){$y_{1}$}
  \put(28.5, 7){$y_{2}$}

  \put(35,9){\line(1,0){4}}
  \put(41,8){$\cdots$}
  \put(47,9){\line(1,0){4}}

  \put(53,16){$i-1$}

  \put(56, 9){\circle{10}}
  \put(53.5, 7){$y_{i-1}$}
  \put(53.5, 10){$y_{i-2}$}


  \put(61,9){\line(1,0){20}}

  \put(67, 12){\vector(1,0){8.5}}
  \put(63.5, 15){$\alpha_{i-1}(y_{i-1})$}

  \put(75.5, 6){\vector(-1,0){8.5}}
  \put(63.5, 0.5){$\beta_{i-1}(y_{i-1})$}

  \put(85,16){$i$}

  \put(86, 9){\circle{10}}
  \put(84.5, 7){$y_{i}$}
  \put(83.5, 10){$y_{i-1}$}


  \put(91,9){\line(1,0){16}}

  \put(94.5, 12){\vector(1,0){8.5}}
  \put(94, 15){$\alpha_{i}(y_i)$}

  \put(103, 6){\vector(-1,0){8.5}}
  \put(94, 0.5){$\beta_{i}(y_i)$}

  \put(108,16){$i+1$}

  \put(112, 9){\circle{10}}
  \put(109.5, 7){$y_{i+1}$}
  \put(110.5, 10){$y_{i}$}

  \put(117,9){\line(1,0){4}}
  \put(123,8){$\cdots$}
  \put(129,9){\line(1,0){4}}

  \put(134,16){$T-1$}

  \put(138, 9){\circle{10}}
  \put(135.5, 7){$y_{T-2}$}
  \put(135.5, 10){$y_{T-1}$}

  \put(143,9){\line(1,0){19}}

  \put(149, 12){\vector(1,0){8.5}}
  \put(145.5, 15){$\alpha_{T-1}(y_{T-1})$}

  \put(157.5, 6){\vector(-1,0){8.5}}
  \put(145.5, 0.5){$\beta_{T-1}(y_{T-1})$}

  \put(166,16){$T$}

  \put(167, 9){\circle{10}}
  \put(164.5, 10){$y_{T-1}$}
  \put(166.5, 7){$y_{T}$}

\end{picture}

\center \caption{Forward-backward computation scheme.}
\end{figure*}

\section{The forward-backward algorithm over a commutative semiring}

%
\begin{definition}
A \emph{commutative semiring} is a tuple $\big(\mathbb K, \oplus,
\otimes, 0, 1 \big)$ where $\mathbb K$ is the set with operations
$\oplus$ and $\otimes$ such that both $\oplus$ and $\otimes$ are
commutative and associative and have identity elements in $\mathbb
K$ ($0$ and $1$ respectively), and $\otimes$ is distributive over
$\oplus$.
\end{definition}

Let $\big(\mathbb K, \oplus, \otimes, 0, 1 \big)$ be a commutative
semiring and let $\bs y=\big\{y_0, \dots, y_T\big\}$ be a set of
variables taking values from the set $\mc Y$ of cardinality $N$.
We define the \textit{local kernel} functions $u_t: \mc Y^2
\rightarrow \mathbb K$ for $t=1, \dots , T$, and the
\textit{global kernel} function $u: \mc Y^{T+1} \rightarrow
\mathbb K$, assuming that the following factorization holds
\begin{equation}
\label{u-chain}%
u(\bs y)=\bigotimes_{i=1}^{T} u_i(y_{i-1}, y_{i})
\end{equation}
for all $\bs y = (y_0, \dots, y_T)  \in \mc Y^{T+1}$.

The \textit{FB} algorithm \cite{Kschischang: FG and SPA},
\cite{Aji-McEliece: GDL} solves two problems

\begin{enumerate}

\item The \textit{marginalization problem:} Computes the sum
\begin{equation}
\label{fsr: v_ab} v_t(y_t, y_{y+1})=%
\bigoplus_{y_{\{k-1, k\}^c}} u(\bs y)=%
\bigoplus_{y_{\{k-1, k\}^c}}%
\bigotimes_{i=1}^{T} u_i(y_{i-1}, y_{i}),
\end{equation}

\item The \textit{normalization problem:} Computes the sum
\begin{equation}
\label{fsr: Z}%
Z = \bigoplus_{\bs y} u(\bs y)=%
\bigoplus_{\bs y}%
\bigotimes_{i=1}^{T} u_i(y_{i-1}, y_{i}).
\end{equation}

\end{enumerate}

The \textit{FB} recursively computes the \textit{forward vector}
\begin{equation}
\label{fsr: alpha_def}
\alpha_i(y_i)=\bigoplus_{y_{0:i-1}}%
\bigotimes_{t=1}^{i} u_t(y_{t-1}, y_{t}),
\end{equation}
which is initialized to
\begin{equation}
\label{FAI} \alpha_0(y_0)=1,
\end{equation}
and recursively computed using
\begin{equation}
\label{FAR}%
\alpha_i(y_i)=\bigoplus_{y_{i-1}}
u_{i-1}(y_{i-1}, y_i) \otimes \alpha_{i-1}(y_{i-1})
\end{equation}%
and the \textit{backward vector}
\begin{equation}
\label{fsr: beta_def}
\beta_i(y_i)=\bigoplus_{s_{i+1:T}}\ %
\bigotimes_{t=i+1}^{T} u_t(y_{t-1}, y_{t}),
\end{equation}
which is recursively computed using
\begin{equation}
\label{BAR}%
\beta_i(y_i)=\bigoplus_{y_{i+1}} 
u_{i+1}(y_i, y_{i+1}) \otimes \beta_{i+1}(y_{i+1})
\end{equation}
and initialized to
\begin{equation}
\label{BAI} \beta_{T}(y_{T})=1.
\end{equation}
Once the forward $\alpha_{k-1}$ and backward $\beta_k$ vectors
are computed, we can solve the marginalization problem by use of
the formula
\begin{equation}
\label{FB MPF}%
\bigoplus_{y_{\{k-1, k\}^c}}
\bigotimes_{i=1}^T u_i(y_{i-1}, y_i) =%
\alpha_{k-1}(y_{k-1}) \otimes u_k(y_{k-1}, y_k) \otimes
\beta_k(y_k)
\end{equation}

The normalization problem can be solved with the forward pass only
according to
\begin{equation}
\label{FB Norm Final}%
\bigoplus_{\bs y}%
\bigotimes_{i=1}^T u_i(y_{i-1},y_i) =%
\bigoplus_{y_{T}} \alpha_{T}(y_{T}).
\end{equation}

\section{Linear-Chain CRF Training using the Forward Backward Algorithm}

Linear-chain CRFs are discriminative probabilistic models over
observation sequences $\bs x = (x_1, \dots, x_T)$ and label
sequences $\bs y = (y_1, \dots, y_T)$, defined with conditional
probability
\begin{equation}
\label{CRF}%
\p=%
\frac{1}{\Z} \prod_{i=1}^T%
\factor{i}.
\end{equation}
The symbol $\langle \cdot,\cdot \rangle$ denotes the scalar
product between an
$M$-dimensional parameter vector %
\begin{equation}
\bs{\theta}=\big[\theta_1 , \dots , \theta_M \big]
\end{equation}
and the \emph{feature vector} on position $i$
\begin{equation}
\featvec{i}=\big[ \feat{1}{i} , \dots , \feat{M}{i} \big].
\end{equation}
The normalization factor
\begin{equation}
\label{CRF Z(x)}%
\Z = \sum_{\bs y} \prod_{i=1}^T%
\factor{i}
\end{equation}
is called the \emph{partition function}.

The goal of the \emph{CRF} training is to build up the model
(\ref{CRF}) from the data set $\{(\bs x^{(d)}, \bs y^{(d)}
\}_{d=1}^D$. The standard method is to maximize the log likelihood
of (\ref{CRF}):
\begin{equation}
\label{CRF LH}%
\mc L(\bs \theta)= \sum_{d=1}^D%
\log p(\bs y^{(d)} | \bs x^{(d)} ; \bs \theta )
\end{equation}
over the parameter vector $\bs \theta$ for the chosen set of
feature vectors $\featurevec{i}$. The maximum can be found with
several of the gradient methods \cite{Gupta},
\cite{Lafferty_et_al_01}, \cite{Sha_Pereira_03}, \cite{Sutton_08},
\cite{Vishwanathan_et_al_06},
which requires the computation of the gradient $\nabla_{\bs \theta}
\mc L(\bs \theta)$. The gradient can be expressed, according to
(\ref{CRF}) and (\ref{CRF LH}), as:
\begin{multline}
\label{Grad LH}
\nabla_{\bs \theta} \mc L(\bs \theta)= %
\sum_{d=1}^D \sum_{i=1}^{T^{(d)}}%
\bs f (y_{i-1}^{(d)} , y_{i}^{(d)}, \bs x^{(d)}, i)- 
\sum_{d=1}^D%
\frac{\nabla_{\bs \theta} \Z}{\Z},
\end{multline}
where $T^{(d)}$ is the length of the $d$-th observation sequence.

The main problem in the evaluation of the log likelihood gradient
(\ref{Grad LH}) is the computation of the quotient between the partition
function gradient and the partition function. The partition
function gradient can be represented as
\begin{equation}
\nabla_{\bs\theta} \Z = [\partialz{1},\dots,\partialz{M}],
\end{equation}
where $\partialz{m}$ denotes the $m$-th partial derivative, and
can be obtained from (\ref{CRF Z(x)}) after the use of the
Leibniz's product rule:
\begin{equation}
\label{Grad Z}%
\gradz=
\sum_{y_{0:T}} \prod_{i=1}^{T}%
\factor{i}
\sum_{k=1}^T%
\featvec{k}.
\end{equation}
The standard method for the computation of the partition function
and its gradient \cite{Lafferty_et_al_01} is based on the
forward-backward algorithm which is reviewed in the following
section.

\subsection{Sum-product semiring forward-backward algorithm}

\begin{definition}
The sum-product semiring is the tuple $\big(\sR,+\cdot,1,0\big)$,
where $\sR$ is the set of real numbers and the operations defined
in a standard way.
\end{definition}

The partition function (\ref{CRF Z(x)}) can be obtained as the
solution of the normalization problem (\ref{FB Norm Final}) for
factorization:
\begin{equation}
\prod_{i=1}^T \factor{i}.
\end{equation}
as
\begin{equation}
\label{Z solution}%
\Z =\ %
\sum_{y_{0:T}}\ %
\prod_{i=1}^T \factor{i}
\end{equation}

The gradient can be computed using the %
solution for the marginalization problem (\ref{FB MPF}) in the
sum-product semiring. First, we change the sum ordering in
(\ref{Grad Z}) and split the sum over $\bs y$ to $y_{\{k-1,k\}}$
and $y_{\{k-1,k\}^c}$ sums, transforming (\ref{Grad Z}) to:
\begin{align}
\label{Grad Z SPA}%
\gradz=\sum_{k=1}^{T}\sum_{y_{\{k-1,k\}}}
\Big(\sum_{y_{\{k-1,k\}^c}}
\prod_{i=1}^{T}%
\factor{i} \Big)\nonumber\\%
\cdot \featvec{k}. 
\end{align}
The marginal values,
\begin{equation}
\label{sp fb marginal}%
\sum_{y_{\{k-1,k\}^c}}
\prod_{i=1}^{T}%
\factor{i},
\end{equation}
as the marginalization problem over the sum-product semiring, can be found by
recursive computation of forward vectors,
\begin{equation}
\alpha_i(y_i)=\sum_{y_{0:i-1}}%
\prod_{t=1}^{i} \factor{t}
\end{equation}
and backward vectors
\begin{equation}
\beta_i(y_i)=\sum_{y_{i+1:T}}\ %
\prod_{t=i+1}^{T} \factor{t}.
\end{equation}


The $FB$ algorithm over the sum-product semiring suffers from
numerical instability since the exponential terms can fall out of
the machine precision scope and it is usually replaced with a more
stable $FB$ algorithm over the log-domain sum-product semiring.


\subsection{Log-domain sum-product semiring forward-backward algorithm}

\begin{definition}
The log-domain sum-product semiring is the tuple $\big(\sR^*,
\oplus, \otimes, -\infty, 0 \big)$ where $\sR^*$ is the extended set of real
numbers and the operations are defined by
\begin{align}
\label{log sp plus}
a \oplus &b = \log(\exp^a + \exp^b)\\
\label{log sp times} &a \otimes b = a + b,
\end{align}
for all $a,b \in \sR$.
\end{definition}

The following lemma follows straightforwardly from the definition
of the log-domain sum-product semiring.
\begin{lemma}
\label{lemma log sp} Let $a_i\in \sR$ for all $1 \leq i \leq T$.
Then, the following equalities hold for log-domain sum-product semiring:
\begin{equation}
\log\big( \sum_{i=1}^T a_i \big)=\bigoplus_{i=1}^{T} \log a_i,\quad %
\log \big( \prod_{i=1}^T a_i \big) = \bigotimes_{i=1}^{T} \log a_i.
\end{equation}
\end{lemma}

In log-domain the local kernels have the form:
\begin{equation}
\label{log sp u_i}%
u_i(y_{i-1},y_i)=\dotprod{i},
\end{equation}%
for $i=1,\dots,T$. According to Lemma \ref{lemma log sp} and
expression (\ref{fsr: alpha_def}), the forward vector in
the log-domain sum-product semiring is the logarithm of the forward vector
in the sum-product semiring:
\begin{align}
\label{log fsr: alpha_def}
\alpha_i(y_i)&=
\logsum{y_{0:i-1}}{}%
\logprod{t=1}{i} \umat{t}=\nonumber\\%
&=\log \Big(\sum_{y_{0:i-1}}%
\prod_{t=1}^{i} \factor{t}\Big).
\end{align}
The forward vector $\alpha_0$ is initialized to $0$ which is the
identity for $\logt$,
\begin{equation}
\label{log FAI} \alpha_0(y_0)=0,
\end{equation}
and it is recursively computed using
\begin{equation}
\label{log FAR}%
\alpha_i(y_i)=\bigoplus_{y_{i-1}}
\big(u_{i}(y_{i-1}, y_{i}) + \alpha_{i-1}(y_{i-1})\big).
\end{equation}%
Similarly to Lemma \ref{lemma log sp} and to expression
(\ref{fsr: beta_def}), the backward vector in the log-domain
sum-product semiring is the logarithm of the backward vector in
sum-product semiring:
\begin{align}
\label{log fsr: beta_def}
\beta_i(y_i)&=%
\logsum{y_{i+1:T}}{}\ %
\logprod{t=i+1}{T} \factor{t}=\nonumber\\%
&=\log \sum_{y_{i+1:T}}\ %
\prod_{t=i+1}^{T} \factor{t},
\end{align}
being initialized to
\begin{equation}
\label{log BAI} \beta_{T}(y_{T})=0,
\end{equation}
and recursively computed using
\begin{equation}
\label{log BAR}%
\beta_i(y_i)=\bigoplus_{y_{i+1}} 
\big(u_{i+1}(y_i, y_{i+1}) + \beta_{i+1}(y_{i+1})\big).
\end{equation}
If the log-domain addition is performed using the definition, $a
\oplus b = \log(\exp^a + \exp^b)$, the numerical precision is
being lost when computing $\exp^a$ and $\exp^b$. But, as noted in
(\cite{Sutton_08}), $\oplus$ can be computed as
\begin{equation}
a \oplus b = a + \log\big(1 + \exp^{(b-a)}\big) =%
b + \log\big(1 + \exp^{(a-b)}\big),
\end{equation}
which can be much more numerically stable, particularly if we pick
the version of the identity with the smaller exponent.

The logarithm of the normalization constant (\ref{Z solution}) is
according to Lemma \ref{lemma log sp}
\begin{equation}
\log \Z = \log \totsum \prod_{i=1}^T \factor{i} =%
\logtotsum \logprod{i=1}{T} \umat{i}, 
\end{equation}
and it can be computed using the solution of normalization problem
in the log-domain sum-product semiring with forward algorithm according
to (\ref{FB Norm Final})
\begin{equation}
\log \Z =  
\logsum{y_{T}}{} \alpha_{T}(y_{T}).
\end{equation}
According to Lemma \ref{lemma log sp}, the marginal values
(\ref{sp fb marginal}) in the log-domain sum-product semiring have the
form
\begin{align}
v_{k}(y_{k-1}, y_k)&=%
\log \sum_{y_{\{k-1,k\}^c}}
\prod_{i=1}^{T}%
\exp^{\dotprod{i}}=\nonumber\\%
&=\logsum{y_{\{k-1,k\}^c}}{} \logprod{i=1}{T} \umat{i}.%
\end{align}
The marginal values can efficiently be computed according to the
solution of the marginalization problems (\ref{FB MPF}):
\begin{equation}
v_{k}(y_{k-1}, y_k)=%
\alpha_{k-1}(y_{k-1}) \otimes u_k(y_{k-1}, y_k) \otimes
\beta_k(y_k),
\end{equation}
where $\alpha_{k-1}(y_{k-1})$ and $\beta_k(y_k)$ are computed with
the $FB$ algorithm over the log-domain sum-product semiring, using
equations (\ref{log FAI})-(\ref{log BAR}).
Then, by taking the logarithm of the $m$-th component in gradient expression (\ref{Grad Z SPA}), we get%
\begin{align}
\label{logfb: ln grad z}
\log & \partialz{m} = \nonumber\\ 
&=\logsum{k=1}{T}\logsum{y_{\{k-1,k\}}}{}%
v_{k}(y_{k-1}, y_k)
\logt \log \feat{m}{k}, 
\end{align}
for $m=1,\dots, M$. 
Finally, the quotient between the partition function and its
gradient can be computed according to
\begin{equation}
\frac{\gradz}{\Z}=\exp^{\log \gradz - \log \Z}.
\end{equation}

\begin{algorithm}[!h]
 \LinesNumbered
    \SetKwComment{Comment}{}{}

\BlankLine
  \SetKwData{Left}{left}\SetKwData{This}{this}\SetKwData{Up}{up}
  \SetKwFunction{Union}{Union}\SetKwFunction{FindCompress}{FindCompress}
  \SetKwInOut{Input}{input}\SetKwInOut{Output}{output}
  \Input{$\bs x$, \ $\bs \theta$, \ $\featvec{k}; \ y_{k-1},y_k \in \mc Y,\ k = 1, \dots ,T; $}
  \Output{$\nabla_{\bs\theta}\Z / \Z ;$ }
  \BlankLine
  \Comment{ /* Matrices initialization */}
  \BlankLine
  \For{$k\leftarrow 1$ \KwTo $T$}{
    \ForEach{$y_{k-1}$ in $\mc{Y}$}{
    \ForEach{$y_k$ in $\mc Y$}{
           $\umat{k}=\dotprodexp{k}$
    }
    }
 }

    \BlankLine
    \Comment{/* Forward phase */}
    \BlankLine

    \ForEach{$y_{0}$ in $\mc{Y}$}{$\alpha_0 (y_0) \leftarrow 0$;}
    \For{$k\leftarrow 1$ \KwTo $T$}{

    \ForEach{$y_k$ in $\mc Y$}{$\alpha_k(y_k)\leftarrow\logsum{y_{k-1}}{}\big( \umat{k} +  \alpha_{k-1}(y_{k-1})\big)$ }

    }

\BlankLine
    \Comment{/* Backward phase */}
    \BlankLine

    \ForEach{$y_{T}$ in $\mc{Y}$}
    {$\beta_T(y_T) \leftarrow 0$\;}
     \BlankLine

    \For{$k\leftarrow T-1$ \KwTo $0$}{

    \ForEach{$y_k$ in $\mc Y$}{
         $\beta_k (y_k) \leftarrow \logsum{y_{k+1}}{}\big(u_{k+1}(y_{k}, y_{k+1})+ \beta_{k+1}(y_{k+1})\big)$\;
    }
    }

    \BlankLine
    \Comment{/* Termination */}
    \BlankLine

    $\quad\log \Z = \logsum{y_{T}}{} \alpha_{T}(y_{T})$

    \For{$k\leftarrow 1$ \KwTo $T$}{

    \ForEach{$y_{k-1}$ in $\mc{Y}$}{

    \ForEach{$y_k$ in $\mc Y$}{
        $v=\alpha_{k-1}(y_{k-1}) + u_k(y_{k-1},y_k) + \beta_{k}(y_k)$

        \ForEach{$m$ in $\mc A_k(y_{k-1},y_k)$}{
            $lnf \leftarrow \log \feat{m}{k}$\;
            $\log \nabla_m Z \leftarrow \log \nabla_m Z \logp\ (v+ lnf)$

        }
    }
}
}
        \For{$m \leftarrow 1$ \KwTo $M$} {%
            $\partialz{m}/\Z \leftarrow \exp^{\log \nabla_m Z - \log Z}$
         }

\caption{Log-domain $FB$ algorithm}
\label{alg fb}
\end{algorithm}

\begin{table*}[!t]
\center{
 \begin{tabular}{cllllllllll}
&$\tableskip$& $\oplus$ &$\tableskip$& $+$ &\tableskip& $\times$ &$\tableskip$& $\log$ &$\tableskip$& Mem \\
\hline
\\
$u$ && $-$ && $N^2TA$ && $N^2TA$  && $-$ && $N^2T$ \\
$\alpha$ && $N^2T$ && $N^2T$ && $-$&& $-$ && $NT$  \\
$\beta$ && $N^2T$ && $N^2T$  && $-$&& $-$ && $NT$\\
$v$ && $-$ && $2N^2T$ && $-$ && $-$ && $1$\\
$lnf$ && $-$ && $-$ && $-$ && $N^2TA$ && $1$\\
$\log \Z$ && $N$ && $-$ && $-$ && $-$ && $1$\\
$\log \partialz{m}$ && $N^2TA$ && $N^2TA$ && $-$ && $-$ && $M$\\
Asymptotical  && $N^2TA$ && $2 N^2TA$ && $N^2TA$  &&$N^2TA$ && $N^2T + M$\\
\\
\hline
\end{tabular}
\caption{Time and memory complexity of the log-domain $FB$ algorithm.}
\label{tabela FB}}
\end{table*}

\subsection{Time and Memory Complexity}

The time and memory complexity of the algorithm for the
computation of the partition function and its derivatives by the
\emph{FB} algorithm is given in Table 1. The time complexity is
defined as the number of operations required for the execution of
the algorithm for a given pseudo code. In our analysis we consider
real operations (addition and multiplication), log-domain
operations (recall that log-domain multiplication is defined as
real addition) and the number of computed logarithms. The memory
complexity is defined as the number of 32-bit registers needed to
store variables during algorithm execution. The complexity
expressions are simplified by taking the quantities in expressions
to tend to infinity, and keeping only the leading terms. In
discussion, we will use big $\mc O$ notation
\cite{Cormen_et_al_03}.


In applications, the feature functions $\featvec{k}$ map the input
space for a fixed sequence $\bs x$ into sparse vectors, which has
nonzero values only at positions
\begin{equation}
\mc A_k(y_{k-1}, y_k)=\big\{\ m \ ; \ \feat{m}{k} \text{ is
nonzero}\ \big\},
\end{equation}
which allows the complexity reduction by performing the
computation only for nonzero elements. In our analysis we
will use the average number of nonzero elements defined as
\begin{equation}
A = \frac{\sum_{k=1}^T \mc A_k(y_{k-1}, y_k)}{T}.
\end{equation}

As Table 1 shows, the computationally most demanding part of the
algorithm is the termination phase, which requires $\mc O(N^2 T
A)$ log-additions (recall that one log-addition requires the
computation of the exponent and logarithm). The memory complexity
of the algorithm is $\mc O(N^2T+M)$, governed by the space needed
for storing the matrices $u_i$.
%
%
%
The dependence of the memory complexity on the sequence length can
significantly decrease computational performances of the algorithm
if a long sequence is used, since it can cause overflows from the
internal system memory to the disk storage, as shown in section
\ref{sec: ex}.

The memory complexity can be reduced by the re-computation of the
matrices $u_i$ in the backward pass (line 14) and in the termination step
(line 19), but this leads to the increased total number of
additions and multiplications for $2N^2TA$, 
while the memory complexity still depends on the sequence length
since all forward and backward vectors need to be stored. The
further improvement can be achieved if one notes that the backward
vectors are computed during the termination step since they are
used only once in line 19. In this case, each backward vector can
be deleted after use in line 19 and all backward vectors can
be stored at the memory location not depending on the sequence
length. Then, the matrices $u_i$ can be recomputed only once in
the termination step, where they are used for the computation of
the backward vector 
but, again, all forward vectors need to be stored and the memory
complexity is $\mc O(NT+M)$, still depending on the sequence length.

The problem of memory complexity of the forward-backward algorithm
for the \emph{HMM} has already been studied by Khreich et al. in
\cite{Khreich_et_al_10}. In this paper they have proposed the
algorithm for the computation of marginal probabilities called
forward filtering backward smoothing (\emph{EFFBS}), which runs
with the memory complexity independent of the sequence length,
$\mc O(N)$, with the same asymptotical computational complexity as
the standard forward-backward algorithm. However, the algorithm is
based on the \emph{HMM} assumption that the transition matrix is
constant, and as such cannot be applied to \emph{CRF}s. Khreich
et al. also gave a good review of the previously developed
techniques for memory reduction such as checkpointing and
forward-only algorithm, which try to reduce the memory complexity
of the \emph{FB} algorithm at the cost of computational overhead,
and these techniques can be modified to deal with \emph{CRF}s.

The checkpointing algorithm \cite{Grice_et_al_97},
\cite{Tarnas_et_al_98} divides the input sequence into $\sqrt T$
and during the forward pass only stores the first forward vector in
each sub-sequence (checkpoint vectors). In the backward pass, the
forward values for each sub-sequence are sequentially recomputed,
beginning with checkpoint vectors. In this way, the computational
complexity required for the computation of the forward and backward
vectors is increased to $\mc O (2T-N^2 \sqrt T )$, while the
matrices $u_i$ should also be recomputed, which leads to greater
total computational cost. On the other hand, the memory
complexity, although reduced to $\mc O(N\sqrt T )$, still depends
on the sequence length.

In the forward-only algorithm \cite{Churbanov_Winters-Hilt_08},
\cite{Khreich_et_al_10}, \cite{Miklos_Meyer_05},
\cite{Sivaprakasam_Shanmugan_95}, the expression of the form is
obtained from three-dimensional matrices which are recursively
computed. For the $HMM$, the computation can be realized in the
constant memory space independent of the sequence length $\mc O
(N^2+N)$ and with time complexity $\mc O(N^4T)$. However, if it is
applied to the $CRF$, its time complexity increases to $\mc O(N^4 M
T)$, which is significantly slower than the $FB$ algorithm.

%

In the following section we derive a forward-only algorithm which
operates with the time complexity of order $\mc O(N^2 (M+A) T)$,
while keeping the memory complexity independent of the sequence
length.


\section{Log-domain expectation semiring forward algorithm}
\begin{figure*}[!t]
 \centering
\setlength{\unitlength}{1mm}
\begin{picture}(170, 15)

  \put(4,16){$1$}

  \put(5, 9){\circle{10}}
  \put(3.5, 10){$y_{0}$}
  \put(3.5, 7){$y_{1}$}


  \put(10,9){\line(1,0){15}}

  \put(13, 12){\vector(1,0){8.5}}
  \put(12.5, 15){$\alpha_{1}^{(z)}(y_1)$}

  \put(13, 6){\vector(1,0){8.5}}
  \put(12.5, 0.5){$\alpha^{(h)}_{1}(y_1)$}

  \put(29,16){$2$}

  \put(30, 9){\circle{10}}
  \put(28.5, 10){$y_{1}$}
  \put(28.5, 7){$y_{2}$}

  \put(35,9){\line(1,0){4}}
  \put(41,8){$\cdots$}
  \put(47,9){\line(1,0){4}}

  \put(53,16){$i-1$}

  \put(56, 9){\circle{10}}
  \put(53.5, 7){$y_{i-1}$}
  \put(53.5, 10){$y_{i-2}$}


  \put(61,9){\line(1,0){20}}

  \put(67, 12){\vector(1,0){8.5}}
  \put(63.5, 15){$\alpha_{i-1}^{(z)}(y_{i-1})$}

  \put(67, 6){\vector(1,0){8.5}}
  \put(63.5, 0.5){$\alpha^{(h)}_{i-1}(y_{i-1})$}

  \put(85,16){$i$}

  \put(86, 9){\circle{10}}
  \put(84.5, 7){$y_{i}$}
  \put(83.5, 10){$y_{i-1}$}


  \put(91,9){\line(1,0){16}}

  \put(94.5, 12){\vector(1,0){8.5}}
  \put(94, 15){$\alpha_{i}^{(z)}(y_i)$}

  \put(94.5, 6){\vector(1,0){8.5}}
  \put(94, 0.5){$\alpha^{(h)}_{i}(y_i)$}

  \put(108,16){$i+1$}

  \put(112, 9){\circle{10}}
  \put(109.5, 7){$y_{i+1}$}
  \put(110.5, 10){$y_{i}$}

  \put(117,9){\line(1,0){4}}
  \put(123,8){$\cdots$}
  \put(129,9){\line(1,0){4}}

  \put(134,16){$T-1$}

  \put(138, 9){\circle{10}}
  \put(135.5, 7){$y_{T-2}$}
  \put(135.5, 10){$y_{T-1}$}

  \put(143,9){\line(1,0){19}}

  \put(149, 12){\vector(1,0){8.5}}
  \put(145.5, 15){$\alpha_{T-1}^{(z)}(y_{T-1})$}

  \put(149, 6){\vector(1,0){8.5}}
  \put(145.5, 0.5){$\alpha_{T-1}^{(h)}(y_{T-1})$}

  \put(166,16){$T$}

  \put(167, 9){\circle{10}}
  \put(164.5, 10){$y_{T-1}$}
  \put(166.5, 7){$y_{T}$}

\end{picture}
\center \caption{$EMP$ computation scheme.}
\label{EMP_chain}
\end{figure*}
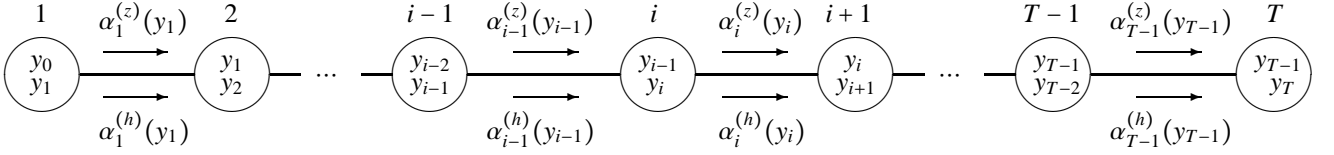

In this section we consider a memory-efficient algorithm for
$CRF$ gradient computation which 
operates as a $FB$ algorithm over an expectation semiring and
develop its numerically stable log-domain version.
In our previous work \cite{Ilic_et_al}, we have developed the
Entropy Message Passing $EMP$, which operates as a forward algorithm
over the entropy semiring, which is the special case the
expectation semiring. Although the algorithms presented in this
paper are more general, in the following text they will be called the
$EMP$, since they operate in the same manner as the algorithm
from \cite{Ilic_et_al}.

\subsection{Expectation semiring forward algorithm}

\begin{definition}
\label{EMP1 ESR def} The expectation semiring of an order $M$ is a
tuple $\left\langle \ \sR \times \sR^M,\ \esrp,\ \esrt, \ ( 0
,\ \bs 0 ), ( 1 ,\ \bs 0)\ \right\rangle$, where the operations
$\esrp$ and $\esrt$ are defined with:
\begin{align}
\label{ESR addition}
(z_1, \bs h_1) \esrp (z_2, \bs h_2)  &= (z_1 + z_2, \bs h_1+ \bs h_2), \\
(z_1, \bs h_1) \esrt (z_2, \bs h_2) &= %
(z_1 z_2 ,\ z_1 \bs h_2 + z_2 \bs h_1),
\end{align}
for all $(z_1, \bs h_1)$, $(z_2,  \bs h_2)$ from $\sR \times
\sR^M$, and $\bs 0$ denotes zero vector. The
first component of an ordered pair is called a $z$-part, while the
second one is an $h$-part.
\end{definition}
For $M=1$, the expectation semiring reduces to the entropy semiring
considered in \cite{Ilic_et_al}. According to the addition rule,
the $z$ and $h$ components of sum of two pairs are the sums of $z$
and $h$ components respectively, which gives us the following lemma.
\begin{lemma}
\label{lemma esr addition} Let $(z_i, \bs h_i)\in \sR \times
\sR^M$ for all $1 \leq i \leq T$. Then, the following equality holds in
the expectation semiring:
\begin{equation}
\label{esr sum operator}%
\esrsum{i=1}{T}(z_i,z_i \bs h_i)=%
\biggl(\ \sum_{i=1}^{T} z_i \ ,%
\ \sum_{i=1}^{T} \bs h_i \ \biggl).\\
\end{equation}
\end{lemma}
Note that if the pairs have the form $(z,z \bs h)$, the multiplication acts
as
\begin{equation}
(z_1, \bs h_1) \esrt (z_2, \bs h_2) = %
(z_1 z_2 ,\ z_1 z_2 (\bs h_1 + \bs h_2).
\end{equation}
This can be generalized with the following lemma.
%
%
\begin{lemma}
\label{lemma esr multiplication} Let $(z_i, z_i \bs h_i)\in \sR
\times \sR^M$ for all $1 \leq i \leq T$. Then, the following
equality holds in the expectation semiring:
\begin{equation}
\label{esr prod operator}%
\esrprod{i=1}{T}(z_i,z_i \bs h_i)=%
\biggl(\ \prod_{i=1}^{T} z_i \ ,%
\ \prod_{i=1}^{T} z_i \ \cdot  \sum_{j=1}^{T} \bs h_j \ \biggl)
\end{equation}
\end{lemma}

According to lemma \ref{lemma esr multiplication}, if the local kernels have the form:
\begin{align}
\label{ESR u_i}%
u_i(y_{i-1},y_i)=%
\big(&\factor{i},\nonumber \\
&\factor{i} \cdot \featvec{i} \big),
\end{align}%
for $i=1,\cdots,T$, the global kernel is, according to the Lemma
\ref{lemma esr multiplication}
\begin{align}
\label{emp: global kernel}
\bigotimes_{i=1}^{T} u_i(y_{i-1}, y_{i})=
\bigl(& \prod_{i=1}^{T} \factor{i},\\ \nonumber%
&\prod_{i=1}^{T}\factor{i} \cdot%
\sum_{j=1}^{T} \featvec{j} \bigl).
\end{align}%
By applying lemma \ref{lemma esr addition} to the expression (\ref{emp: global kernel}), we can obtain the partition function (\ref{CRF Z(x)}),
\begin{equation}
\Z=\totsum \prod_{i=1}^T \factor{i},
\end{equation}
and its gradient (\ref{Grad Z}):
\begin{equation}
\partialz{m}=\totsum \prod_{i=1}^T \factor{i} \cdot \sum_{j=1}^T \featvec{j},
\end{equation}
as $z$ and $h$ parts of the sum
\begin{equation}
\label{u-chain ESR}%
\bigoplus_{\bs y}\bigotimes_{i=1}^{T}%
u_i(y_{i-1}, y_{i})=
\big(\ \Z,%
\ \gradz \big).
\end{equation}

%

The expression (\ref{u-chain ESR}) can be computed as the
normalization problem (\ref{FB Norm Final}) by use of the forward
algorithm over the expectation semiring. Note that $z$-parts of
addition and multiplication acts as addition and multiplication in the
sum-product semiring. Accordingly, the $z$-parts of forward
vectors will be the same as the forward vectors in the sum-product
semiring, and their computation is numerically unstable. In the
following subsection we a develop numerically stable forward
algorithm which operates over a log-domain expectation semiring.


\subsection{Log-domain expectation semiring forward algorithm}

The log-domain expectation semiring is a combination of the log-domain
sum-product semiring and the expectation semiring. It can be
obtained if real addition and multiplication in the definition of
expectation semiring operations are replaced with their log-domain
counterparts.

Before we define the log-domain expectation semiring, we introduce
some usefull notation. Firstly, recall that log-domain addition
and multiplication are defined with
\begin{align}
a  \logp b &= \log(\exp^a + \exp^b)\\
a \logt b & = a+b.
\end{align}
The log-product between the scalar $z \in \sR$ and the vector $\bs
h=(h[1],\dots, h[M]) \in\sR^M$ is defined as the vector $z \logt
h$:
\begin{equation}
z \logt \bs h=z \logt (h[1],\dots, h[M])=(z \logt h[1],\dots, z
\logt h[M]),
\end{equation}
the logarithm of the vector $[h_1, \dots, h_M] \in
\sR^M$ is defined as
\begin{equation}
\log [h_1, \dots, h_M] =[\log h_1, \dots, \log h_M].
\end{equation}
The vector $-\bs \infty$ is defined as a vector all of whose coordinates are $-\infty$.

\begin{definition}
\label{EMP1 ESR def} The log-domain expectation semiring of an
order $M$ is a tuple $\left\langle \ \sR \times \sR^M,\ \esrp,\
\esrt, \ ( -\infty ,\ -\bs \infty ), ( 0 ,\ -\bs \infty)\ \right\rangle$, where the
operations $\esrp$ and $\esrt$ are defined with:
\begin{align}
\label{log ESR addition}
(z_1, \bs h_1) &\esrp (z_2, \bs h_2)  = \big(\ z_1 \logp z_2,\ \bs h_1 \logp \bs h_2\ \big), \\
(z_1, \bs h_1) &\esrt (z_2, \bs h_2) =
\big(\ z_1 \logt z_2 ,\ (z_1 \logt \bs h_2)\ \logp \ (z_2 \logt
\bs h_1) \big),
\end{align}
for all $(z_1, \bs h_1)$, $(z_2, \bs h_2)$ from $\sR \times
\sR^M$. Similar to the expectation semiring, the
first component of an ordered pair is called a $z$-part, while the
second one is an $h$-part.
\end{definition}
%
The following lemma is the log-domain version of Lemma \ref{lemma esr addition}.
\begin{lemma}
\label{lemma log esr sum} Let $(z_i, z_i \bs h_i)\in \sR \times
\sR^M$ for all $1 \leq i \leq T$. Then, the following equality
holds in the log-domain expectation semiring:
\begin{equation}
\label{log esr sum operator}%
\osum_{i=1}^{T}(z_i, \bs h_i)=%
\big(\ \logsum{i=1}{T} z_i \ ,%
\ \logsum{i=1}{T} \bs h_i \ \big),
\end{equation}
where
\begin{equation}
\logsum{i=1}{T} a_i = \log\big( \sum_{i=1}^T \textbf{e}^{a_i}
\big).
\end{equation}
\end{lemma}
Similar to the expectation semiring, if the pairs have the form
$(z,\ z \logt \bs h)$ the multiplication acts as
\begin{align}
(z_1, z_1 \logt \bs h_1) \otimes (z_2, z_2 \logt \bs h_2) &= %
(\ z_1 \logt z_2 ,\ z_1 \logt z_2\logt (\bs h_1 \logp \bs h_2)  ).
\end{align}
The following lemma is the log-domain version of Lemma \ref{lemma
esr multiplication}.
\begin{lemma}
\label{lemma log esr times} Let $(z_i,\ z_i \logt \bs h_i)\in \sR
\times \sR^M$ for all $1 \leq i \leq T$. Then, the following
equality holds in the log-domain expectation semiring:
\begin{equation}
\label{log esr times operator}
\bigotimes_{i=1}^{T}(z_i,\ z_i \logt \bs h_i)=%
\big(\ \logprod{i=1}{T} z_i \ ,%
\ \logprod{i=1}{T} z_i \ \logt  \logsum{j=1}{T} \bs h_j \ \big),
\end{equation}
where
\begin{equation}
\logprod{i=1}{T} a_i = \sum_{i=1}^T {a_i}.%
\end{equation}
\end{lemma}


Let for $i=1,\dots, T$
\begin{equation}
\label{log psi} \psimat{i}=\dotprod{i}.
\end{equation}
Then, the logarithm partition function (\ref{CRF Z(x)}) can be
written as
\begin{equation}
\log \Z=\logtotsum \logprod{i=1}{T} \psimat{i}.
\end{equation}
The logarithm of the $m$-th partial derivative can be written as
\begin{equation}
\log \partialz{m}=
\log \Big( \sum_{y_{0:T}} \prod_{i=1}^{T}%
\factor{i}
\sum_{k=1}^T%
\feat{m}{k}\Big),
\end{equation}
or, using the operations from the log-domain sum-product semiring, 
\begin{equation}
\log \gradz=
\logsum{y_{0:T}}{} \logprod{i=1}{T}%
\psimat{i}\logt
\logsum{k=1}T%
\log \featvec{k}.
\end{equation}
If the local kernels have the form:
\begin{equation}
\label{ESR u_i}%
u_i(y_{i-1},y_i)=%
\big(\psimat{i},\ \ \psimat{i} \logt \log \featvec{i} \big),
\end{equation}%
for $i=1,\cdots,T$, the global kernel is, according to Lemma
\ref{lemma log esr times}
\begin{multline}
\esrprod{i=1}{T} u_i(y_{i-1}, y_{i})=
\bigl(\logprod{i=1}{T} \psimat{i},\\%
\logprod{i=1}{T} \psimat{i} \logt%
\logsum{j=1}{T} \log \featvec{j} \bigl).
\end{multline}%
Furthermore,  Lemma (\ref{lemma log esr sum}) for addition in the
expectation semiring implies that the sum of ordered pairs is the
ordered pair of the sums so the partition function and its
gradient can be found as the $z$ and $h$ part of the sum:
\begin{equation}
\label{log u-chain ESR}%
\esrsum{\bs y}{} \esrprod{i=1}{T}%
u_i(y_{i-1}, y_{i})=
\big(\ \log \Z,%
\ \log \gradz \big).
\end{equation}

Expression (\ref{log u-chain ESR}) can be computed as the
normalization problem (\ref{FB Norm Final}) by use of the forward
algorithm over the log-domain expectation semiring (\emph{log-domain EMP algorithm}). The forward algorithm is %
initialized to the  log-domain expectation semiring identity for
the multiplication:
\begin{equation}
\label{EMP FWI} \alpha_0(y_0)=(0, -\bs \infty),
\end{equation}
for all $y_0 \in \mc Y$. After that, we compute other forward
vectors using the recurrent formula
\begin{equation}
\label{esr_alpha_rec} \alpha_i(y_i)=\esrsum{y_{i-1}}{}
u_i(y_{i-1},y_i) \esrt \alpha_{i-1}(y_{i-1}),
\end{equation}
where the local factors are given with (\ref{ESR u_i}).

According to the rules for the addition and multiplication in the
expectation semiring, the $z$ and $h$ parts of recursive equation
(\ref{esr_alpha_rec}) are:
\begin{align}
\label{EMP FWR z}%
\alpha_{i}^{(z)}(y_i)=
&\logsum{y_{i-1}}{}%
\psimat{i} \logt \alpha_{i-1}^{(z)}(y_{i-1})\\
\label{EMP FWR h} 
\bs \alpha_i^{(h)}(y_i)=
&\logsum{y_{i-1}}{}%
\psimat{i}  \logt \bs \alpha_{i-1}^{(h)}(y_{i-1}) \logp \nonumber\\
&\logsum{y_{i-1}}{} \psimat{i} \logt \alpha_{i-1}^{(z)}(y_{i-1})
\logt \log \featvec{i}
\end{align}
for each $y_i \in \mc Y$, $i=1,\ldots T$. 
Finally, the normalization problem can be solved by summation
\begin{equation}
\esrsum{y_{0:T}}{} \esrprod{i=1}{T} u_i(y_{i-1}, y_{i})=
\esrsum{y_{T}}{} \alpha_T(y_{T}), %
\end{equation}
whose $z$ part is a partition function and the $h$ part is its
gradient:
\begin{equation}
\label{termination}
\log \Z=\logsum{y_T}{} \alpha_{T}^{(z)}(y_T), \quad  %
\log \gradz=\logsum{y_T}{} \bs \alpha_T^{(h)}(y_T).
\end{equation}

Hence, the algorithm consists of two parts: i) \emph{forward
pass}, at which the forward vectors are initialized according to
(\ref{EMP FWI}) and recursively computed by (\ref{EMP FWR
z})-(\ref{EMP FWR h}), and during each step the corresponding
matrix $u_i$ is computed and ii) \emph{termination}, at which the
final summation of the forward algorithm is performed according to
(\ref{termination}) and the partition function and its derivatives
are obtained.

Figure \ref{EMP_chain} describes the \emph{EMP} computation
scheme. Recall that the forward-backward based computation
requires that all forward and backward vectors be computed and
stored until the partition function and the derivatives are
obtained  in the termination step. When the \emph{EMP} is used,
the computation terminates when the last forward vector is
computed by use of the formulas (\ref{EMP FWR z}) and (\ref{EMP
FWR h}). This can be realized in the fixed memory space with the
size independent of
the sequence length since the vectors $\alpha_{i-1}^{(z)}$, $\bs \alpha_{i-1}^{(h)}$ %
and the matrices $\psi_i$ should be computed only once in $i-1$-th
iteration and, after having been used for the computation of
$\alpha_i^{(z)}$ and $\bs \alpha_i^{(h)}$, they can be %
deleted. The pseudo code is given in the table Algorithm 2. Here,
the computation is performed using only two pairs of vectors
$(\hat \alpha^{(z)}, \hat {\bs \alpha}^{(h)})$ and $(\alpha^{(z)},
{\bs \alpha^{(h)}})$. Note that the coordinates of the $h$-parts,
$\hat {\bs \alpha}^{(h)}(y_i)$ and $\bs \alpha^{(h)}(y_i)$, are
vectors which carry the information about the gradient and the
$m$-th components of these vectors are denoted with $\hat
\alpha_{[m]}^{(h)}(y_i)$ and $\alpha_{[m]}^{(h)}(y_i)$.

In comparison to the $FB$ algorithm which needs the memory size of
$\mc O(N^2T+M)$, the $EMP$ has a memory complexity $\mc
O(N^2+NM)$, no longer depending on the sequence length $T$ as in
the $FB$ algorithm. The additional cost is paid in time complexity
which is increased for the term $N^2TM$. This is the consequence
of the non-sparse
computation of the $EMP$ $h$ component in line 13. 
Recall that the $FB$ can be completely sparse implemented and, since $A<<M$ in most
of the application, the $FB$ time complexity is lower. 
However, the sparsity can be reduced using the conditionally
trained hidden Markov model assumption considered in
\cite{Feng_et_al_06}, which we used in our implementation. With the
reduced sparsity, the time complexity of the $EMP$ is decreased
and it becomes closer to the $FB$ algorithm. When long sequences are used,
the $EMP$ becomes dominating since the $FB$ needs to use the external
memory. This assertion is justified in the following section,
where we compare the two algorithms on a real data example.




%

\begin{algorithm}[!h]
\label{EMP algorithm}

\LinesNumbered

\SetKwComment{Comment}{}{}

\BlankLine
  \SetKwData{Left}{left}\SetKwData{This}{this}\SetKwData{Up}{up}
  \SetKwFunction{Union}{Union}\SetKwFunction{FindCompress}{FindCompress}
  \SetKwInOut{Input}{input}\SetKwInOut{Output}{output}
  \Input{$\bs x$, $\bs \theta$, \ $\featvec{j}$; $j = 1, \dots , T, \ y_{j-1},y_j \in \mc Y;$}
  \Output{$\gradz / \Z$ ;}

  \BlankLine
  \Comment{ /* Forward algorithm */}

\BlankLine
    \ForEach{$y_0$ in $\mc Y$}{
    $\alpha^{(z)} (y_0)\leftarrow 1$

        \For{$m\leftarrow 1$ \KwTo $M$}{
        $\alpha_{[m]}^{(h)}(y_0)\leftarrow \bs -\bs \infty;$
        }
    }

    \For{$i\leftarrow 1$ \KwTo $T$}{
    \ForEach{$y_i$ in $\mc Y$}{
        \ForEach{$y_{i-1}$ in $\mc{Y}$}{
            $\psi(y_{i-1}, y_{i})=\dotprod{i}$\;
        }
    }

     \ForEach{$y_i$ in $\mc Y$}{
%
            $\hat{\alpha}^{(z)}(y_i)= \logsum{y_{i-1}}{} (\psi(y_{i-1}, y_{i}) + \alpha^{(z)}(y_{i-1}))$\;
            }


%
%
%

        \For{$m \leftarrow 1$ \KwTo $M$}{
        \ForEach{$y_i$ in $\mc Y$}{
    $\hat{\alpha}_{[m]}^{(h)}(y_i) \leftarrow  \logsum{y_{i-1}}{} (\psi(y_{i-1}, y_{i})  + \alpha_{[m]}^{(h)}(y_{i-1}))$\;
            }
        }

        \ForEach{$y_{i-1}$ in $\mc Y$}{
        \ForEach{$y_i$ in $\mc Y$}{
        $\gamma(y_{i-1}) \leftarrow \psi(y_{i-1}, y_{i}) +
        \alpha^{(z)}(y_{i-1})$\;
                \ForEach{$m$ in $\mc A(y_{i-1},y_i)$}{
            $lnf \leftarrow \log \feat{m}{i}$
            $\hat{\alpha}_{[m]}^{(h)}(y_i) \leftarrow \hat{\alpha}_{[m]}^{(h)}(y_i) \logp (\gamma(y_{i-1}) + lnf)$\;

                }
            }
        }

\ForEach{$y_{i}$ in $\mc{Y}$}{

$\alpha^{(z)}(y_i) \leftarrow \hat{\alpha}^{(z)}(y_i);$

\For{$m \leftarrow 1$ \KwTo $M$} {%
$\alpha_{[m]}^{(h)}(y_i) \leftarrow
\hat{\alpha}_{[m]}^{(h)}(y_i);$ } }

     }

\BlankLine
   \Comment{ /* Termination */}
   \BlankLine

   $\log Z \leftarrow \logsum{y_T}{} \alpha^{(z)}(y_{T})$



%
%
            $\log \nabla_m Z \leftarrow \logsum{y_T}{} \alpha^{(h)}_{[m]}(y_{T})$

        \For{$m \leftarrow 1$ \KwTo $M$} {%
            $\partialz{m}/\Z \leftarrow \exp^{\log \nabla_m Z - \log Z}$
         }

\caption{Log-domain $EMP$ algorithm}
\end{algorithm}


\begin{table*}[!t]
\center{
\begin{tabular}{cllllllllll}
&$\tableskip$& $\oplus$ &$\tableskip$& $+$ &\tableskip& $\times$ &$\tableskip$& $\log$ &$\tableskip$& Mem \\
\hline
\\
$\psi$ && $-$ && $N^2TA$ && $N^2TA$ &&  $-$ && $N^2$ \\
$\hat \alpha^{(z)}$ && $N^2T$ && $N^2T$ && $-$&& $-$ &&  $N$\\
$\gamma$ && $-$ && $N^2T$ && $-$&& $-$ &&  $1$\\
$\hat \alpha_m^{(h)}$ && $N^2T(M+A)$ && $N^2T(M+A)$ &&  $-$ && $N^2TA$ && $NM$\\
$\alpha^{(z)}$ && $-$ && $-$ &&  $-$ && $-$&& $N$\\
$lnf$ && $-$ && $-$ &&  $-$ && $N^2TA$&& $1$\\
$\alpha_r^{(h)}$ && $-$ && $-$ &&$-$ && $-$ &&  $NM$\\
$\log \Z$ && $N$ && $-$ && $-$&& $-$ && $1$\\
$\log \partialz{m}$ && $NM$ && $-$ && $-$&& $-$ &&  $M$\\
Asymptotical && $N^2T(M+A)$ && $N^2T(M+2A)$ && $N^2TA$ && $N^2TA$ && $N^2+NM$\\ \\
\hline
\end{tabular}
\caption{Time and memory complexity of the log-domain $EMP$
algorithm.}%
\label{tabela EMP}}
\end{table*}

\begin{figure*}[!t]
\centering
\includegraphics[width=6.5cm]{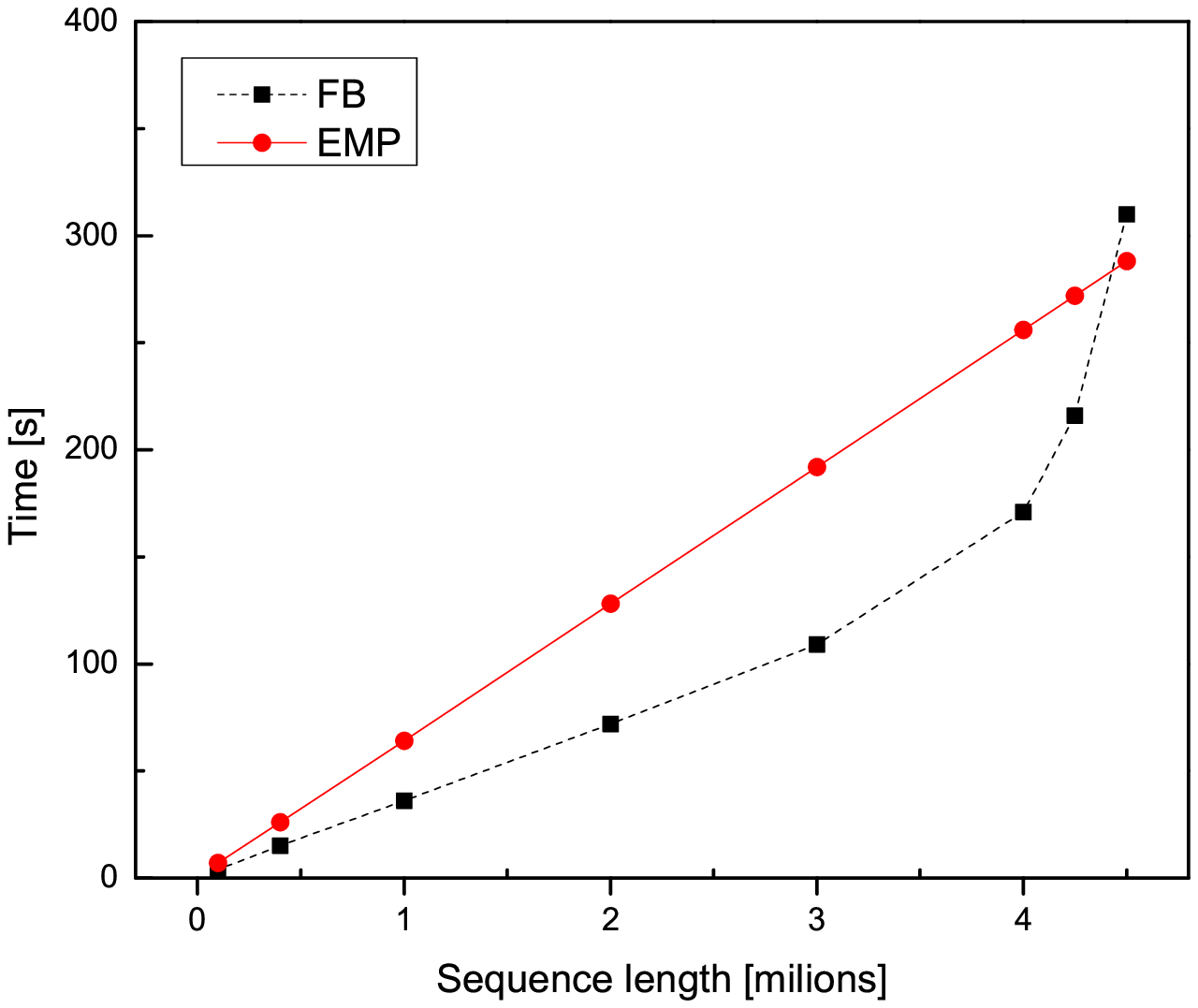}%
\includegraphics[width=6.5cm]{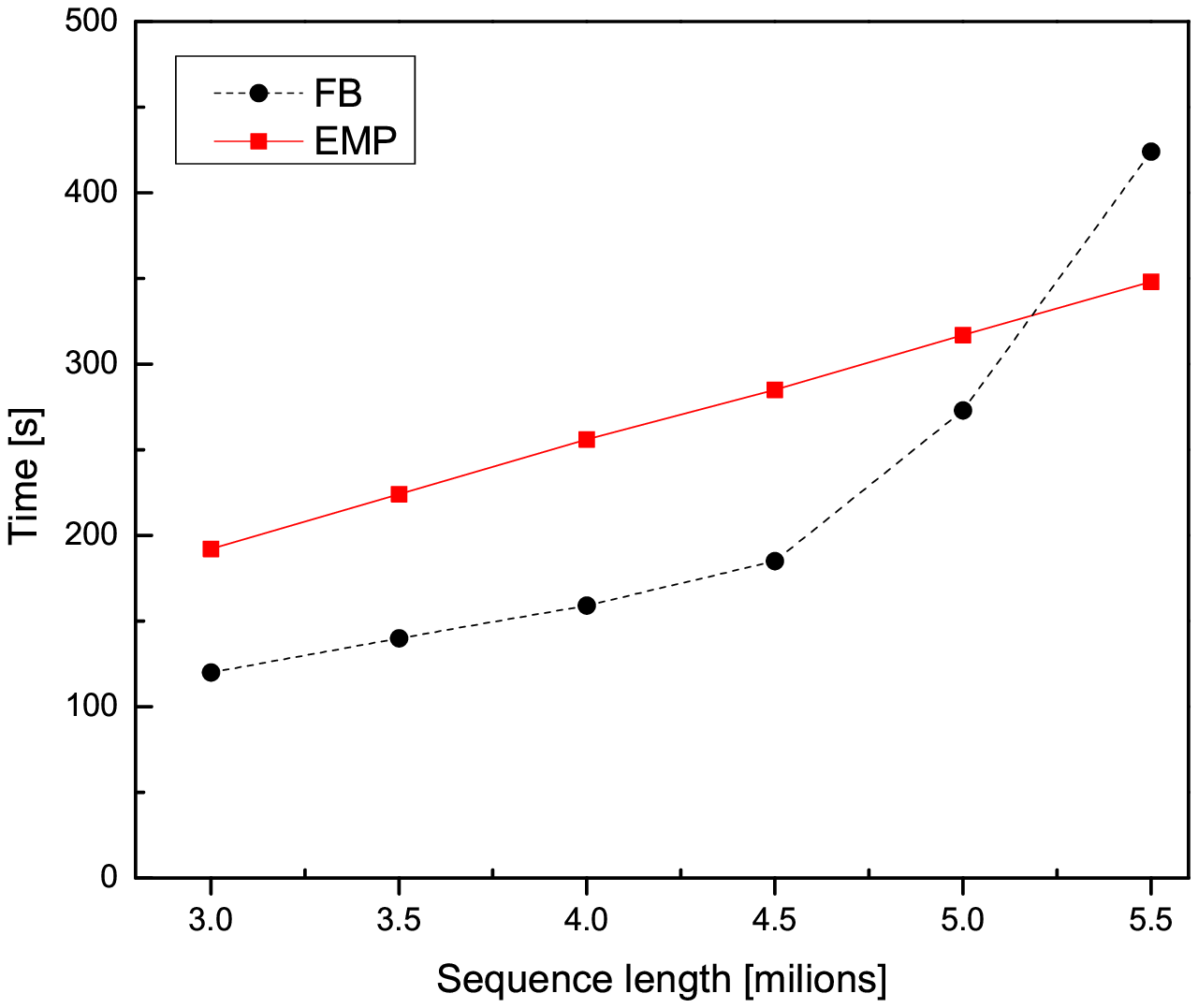}%
\includegraphics[width=6.5cm]{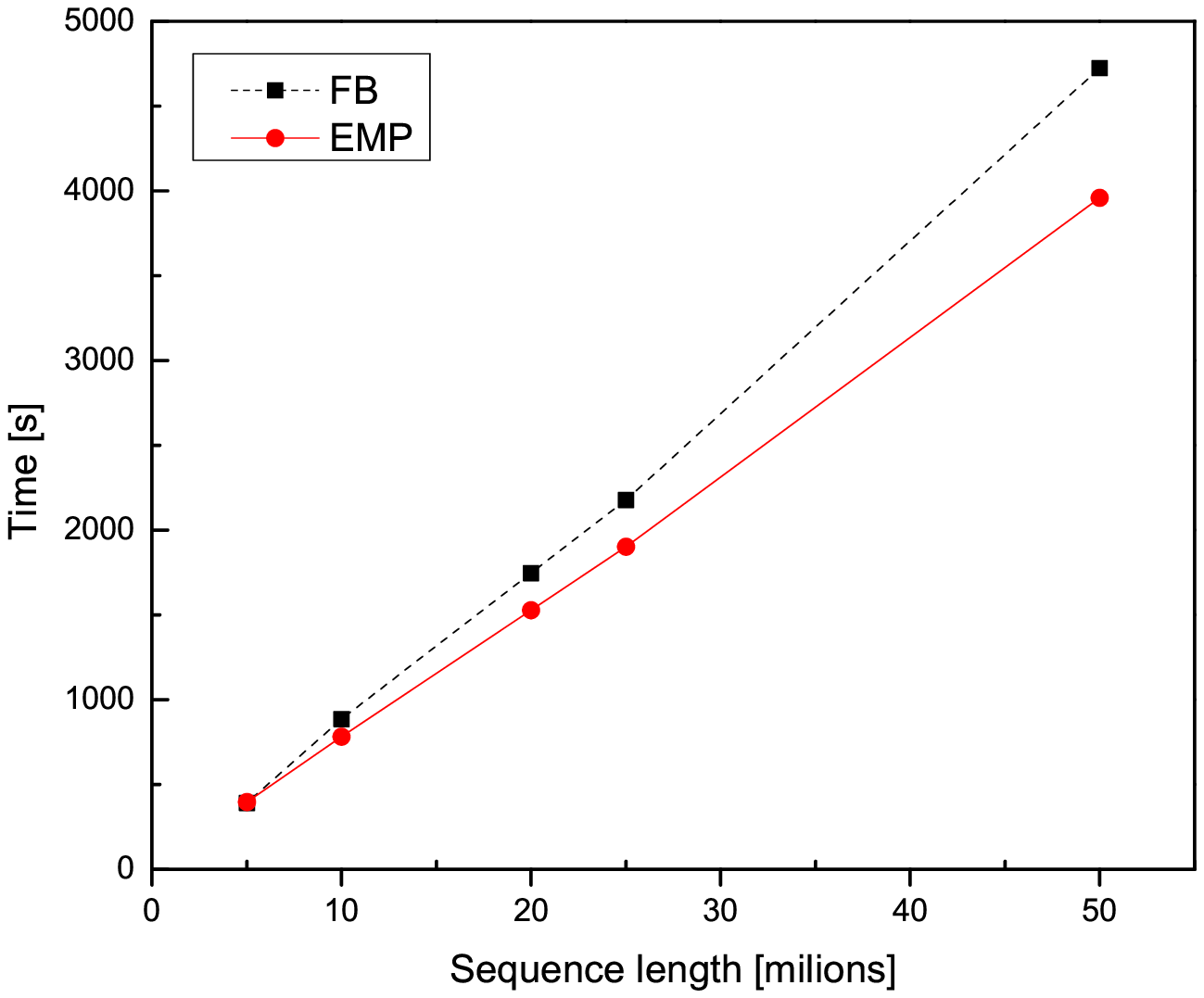}%
\end{figure*}

%
\begin{figure*}[!t]
\centering \subfloat[Case I]%
{\label{fig: FB vs EMP: case I}
\includegraphics[width=6.5cm]{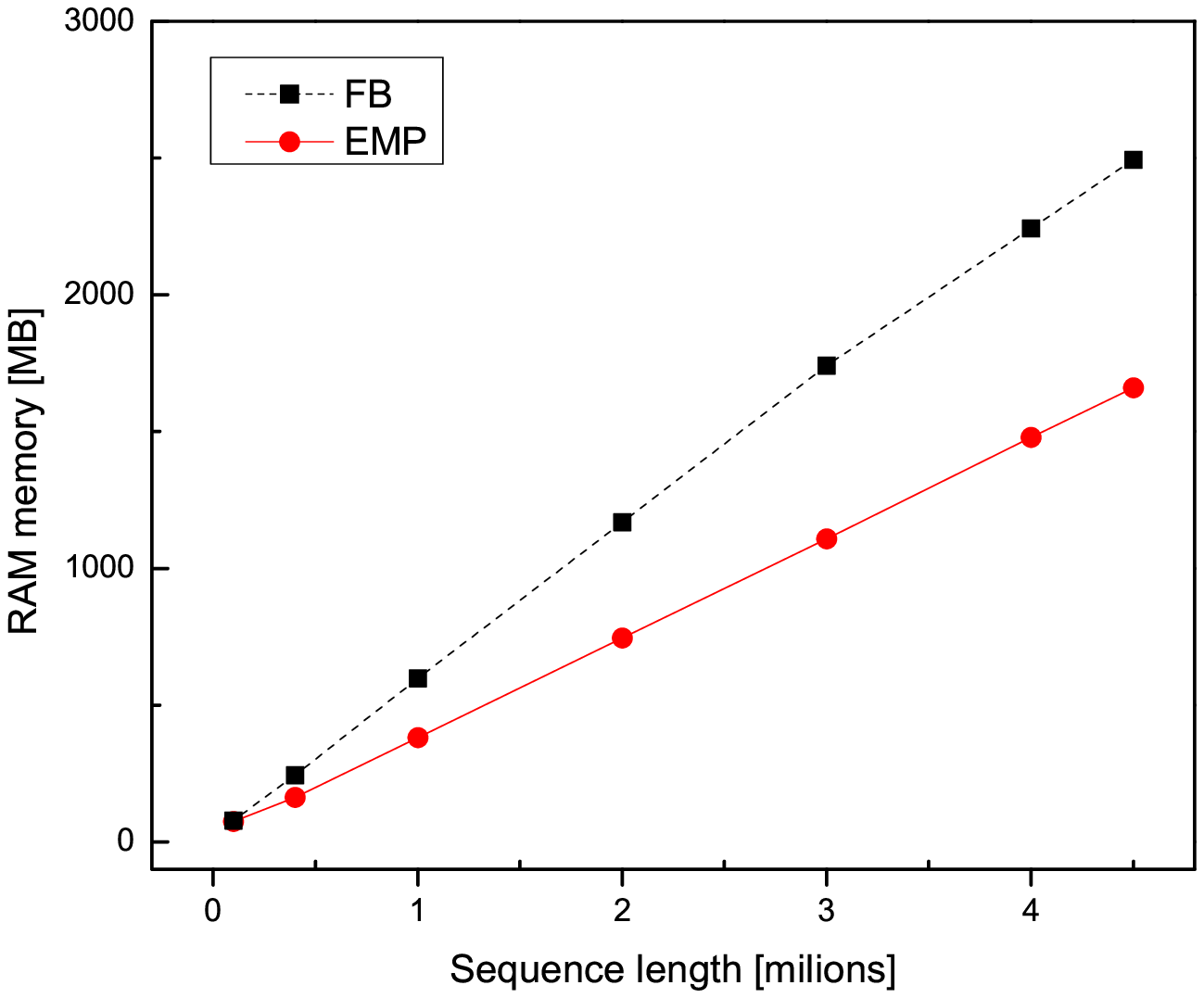}}%
\subfloat[Case II]%
{\label{fig: FB vs EMP: case II}
\includegraphics[width=6.5cm]{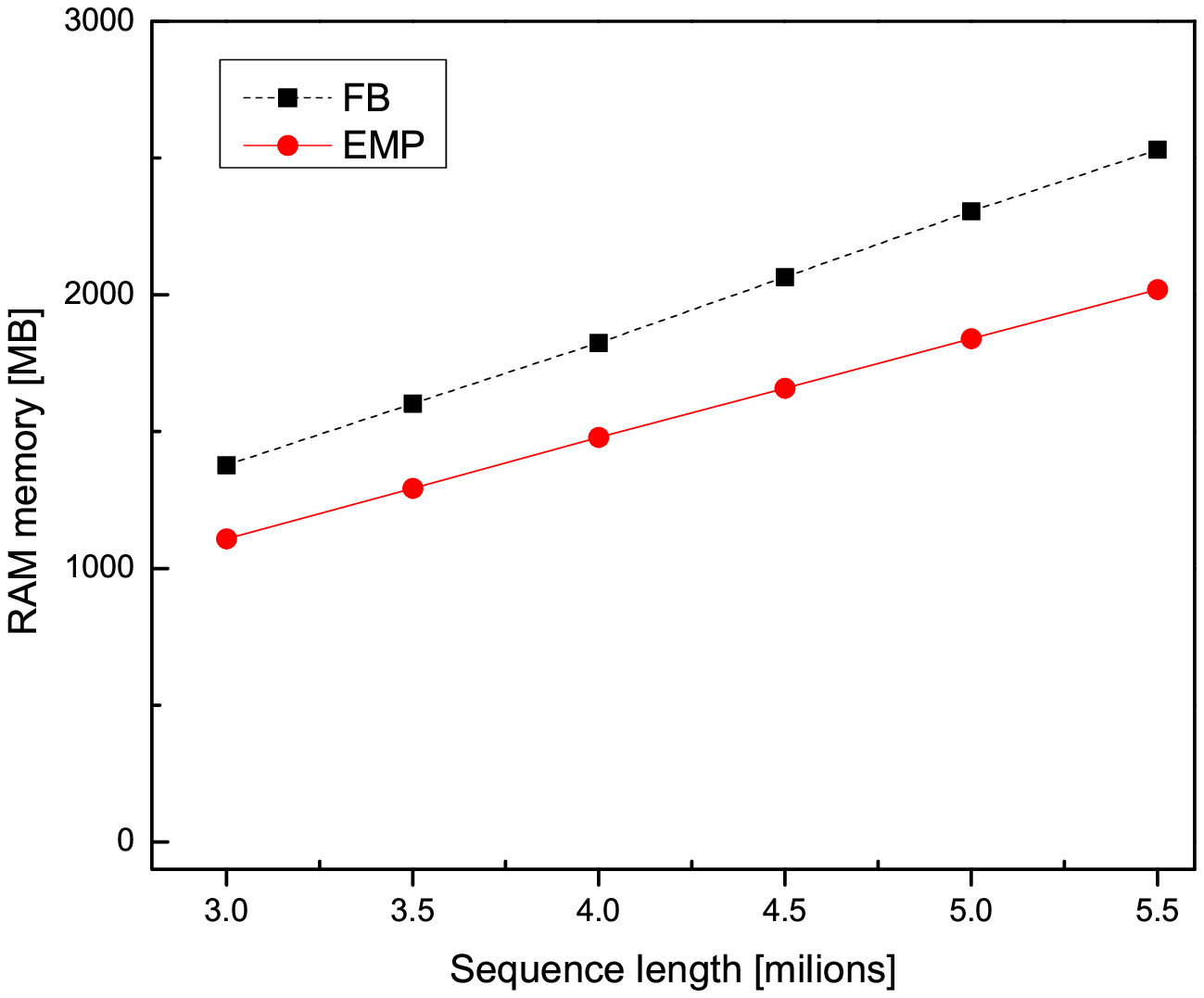}}%
\subfloat[Cases III and IV]%
{\label{fig: FB vs EMP: case III and IV}
\includegraphics[width=6.5cm]{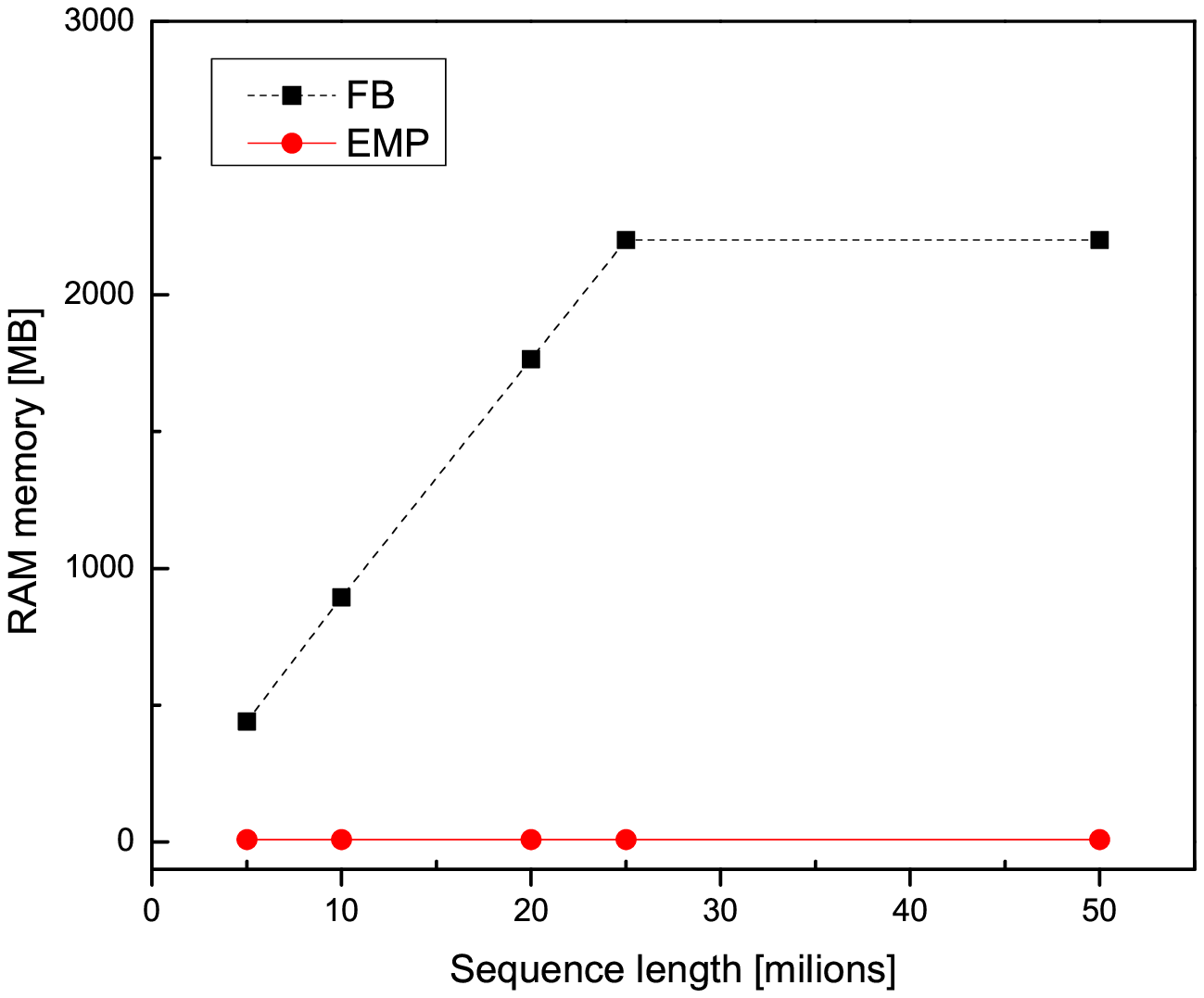}}
\caption{CPU and RAM usage of $FB$ and $EMP$ algorithms}
\label{fig: FB vs EMP}
\end{figure*}

\section{Experiments}
\label{sec: ex}

The intrusion detector learning task is to build a predictive
model capable of distinguishing between "bad" connections, called
intrusions or attacks, and "good" normal connections. Conditional
random fields have proven to be very effective in detecting
intrusion \cite{Gupta_et_al_07}.

As we have already mentioned, in the standard $CRF$ training based on the
$FB$ algorithm the storage requirements are high when long train
sequences are used. This may cause overflows from the internal
system memory to disk storage which decreases computational
performances, since accessing paged memory data on a typical disk
drive is significantly slower than accessing data in RAM
\cite{Khreich_et_al_10},\cite{Warrender_et_al_99}. On the other
hand, the $EMP$ runs with a small fixed memory and it becomes
preferable for long sequences.

In Figure \ref{fig: FB vs EMP} we show the time and memory usage
of both algorithms as functions of the sequence length. The
experiments are performed using a computer with 3GB RAM  and
IntelCore 2 Duo CPU 2.33GHz. In our experiments, we used a KDE corpus
\cite{KDE} for sequences up to $5$ million, while the sequences
longer than $5$ million are created by the concatenation of the KDE
corpus on itself.
%
%
%
%
%
%
%
We consider four different implementation cases depending on the
sequence length:


\textbf{\emph{Case I:}} This corresponds to short sequences, with the
length shorter than $4$ million. This case corresponds to the basic
version of the $FB$ algorithm (Algorithm \ref{alg fb}). In this case, the
sequence is stored in RAM and the RAM usage of both algorithms
linearly grows with the sequence length (Figure \ref{fig: FB vs
EMP: case I}). However, the $FB$ algorithm uses $\mc O(N^2T + M)$
memory for storing intermediate results and its RAM usage grows
faster in comparison to the $EMP$, which needs fixed-size additional
space $\mc O(N^2 + NM)$. RAM usage growth reflects on the computational
performances of $FB$ algorithm, which runs faster then the $EMP$ for
the sequences with the length up to $4.5$ million. As Figure
\ref{fig: FB vs EMP: case I} shows, at the sequence length of $3$
millions $FB$ RAM usage becomes considerable and the $FB$ growth becomes
nonlinear due to the memory paging. Finally, at the sequence
length of about $4.5$ million the $EMP$ becomes faster than $FB$.
One possibility for $FB$ memory reduction is recomputation of
transition matrices which is done in the Case II.

\textbf{\emph{Case II:}} This corresponds to middle length
sequences, between $4$ and $5$ million. At this case the sequence
and all intermediate results are stored in RAM, but the transition
matrices are recomputed every time they are used. Similar to the Case I,
as the sequence becomes longer, the memory required
for storing the forward vectors increases and, for sequences longer
then $5$ million, $FB$ algorithm becomes
slower than the $EMP$ (see Figure \ref{fig: FB vs EMP: case II}).

\textbf{\emph{Case III:}} This corresponds to long sequences
between $5$ and $25$ million. As in Case II, transition matrices
are recomputed and all another intermediate results are stored in
RAM, but the sequence cannot fit in RAM and needs to be stored on
the secondary memory.
%
%
In the $FB$ the sequences have to be read twice from secondary memory, once in the
forward and once in the backward phase. On the other hand, the $EMP$ uses
a single forward pass and reads the sequence only once, which makes
it faster than the $FB$ (Figure \ref{fig: FB vs EMP: case III and
IV}).


\textbf{\emph{Case IV:}} This corresponds to very long sequences
loger than $25$ million. In this case, similar to the Case $III$,
the sequence is stored on the secondary memory. To avoid the $FB$ performance
decreasing due to a large number of intermediate variables stored in
$RAM$, the portion of variables is stored on the secondary memory, which
keeps the RAM usage constant, no longer dependent on the sequence
length (Figure \ref{fig: FB vs EMP: case III and IV}). This
increases the number of accessions to the secondary memory, which further
decreases $FB$ performances in comparison to Case $III$. On the
other hand, the $EMP$ does not need to store additional data on the secondary memory and has the same time growth as in Case $III$, while using a
small constant memory.

The previous results can vary with different operating systems and
used hardware. Nevertheless, the access to secondary memory is very
expensive operation and the algorithm with a low memory complexity
has the advantage, when all data cannot fit in RAM, since the secondary memory accesses can be avoided.


\section{Conclusion}

In this paper, we have developed a numerically stable algorithm
for the computation of the linear-chain CRF gradient. As opposed
to the standard way of finding a CRF gradient by use of the
forward-backward algorithm, the calculation by the proposed
algorithm requires only the forward pass and can be realized with
the memory independent of the observation sequence length. This
makes the algorithm useful in the long sequence labeling tasks
found in computer security \cite{Lane_00},
\cite{Warrender_et_al_99}, bioinformatics
\cite{Krogh_et_al_94},\cite{Mayer_Durbin_04}, and robot navigation
systems \cite{Koening_Simmons_96}.

The proposed algorithm operates as a forward algorithm over the
log-domain expectation semiring, which can be seen as a
modification of the expectation semiring used in the automata
theory and probabilistic context free grammars \cite{Eisner:
Parameter Estimation for PFST}, \cite{Li_Eisner_09}. As mentioned
in the paper, the use of the expectation semiring leads to
numerically unstable algorithms and its log-domain counterpart can
also be applied to numerically stable solutions of problems
considered in \cite{Eisner: Parameter Estimation for PFST},
\cite{Li_Eisner_09}.

\end{document}